%% file: main.tex
\documentclass[10pt,twocolumn,letterpaper]{article}
\usepackage[T2A,T1]{fontenc}
\usepackage[utf8]{inputenc}
\usepackage[russian,english]{babel}

\usepackage{iccv}              


\definecolor{iccvblue}{rgb}{0.21,0.49,0.74}
\usepackage[pagebackref,breaklinks,colorlinks,allcolors=iccvblue]{hyperref}

\usepackage{subfiles}
\usepackage[utf8]{inputenc} 
\usepackage[T1]{fontenc}    
\usepackage{amssymb}
\usepackage{url}            
\usepackage{booktabs}       
\usepackage{amsfonts}       
\usepackage{nicefrac}       
\usepackage{microtype}      
\usepackage[ruled,vlined]{algorithm2e}
\usepackage{multirow}
\usepackage{multicol}
\usepackage{colortbl}
\usepackage{subcaption}
\usepackage{tabularx}
\usepackage{amsmath}
\usepackage[normalem]{ulem}
\useunder{\uline}{\ul}{}
\usepackage{pifont}
\usepackage{makecell}
\usepackage{arydshln}
\usepackage{caption}
\usepackage{adjustbox}
\usepackage{amssymb}
\usepackage{pifont}
\newcommand{\cmark}{\text{\ding{51}}}
\newcommand{\xmark}{\text{\ding{55}}}
\usepackage{array}
\usepackage{wrapfig}
\usepackage{float}

\definecolor{palegreen}{HTML}{dcf6dd}

\def\paperID{10859} 
\def\confName{ICCV}
\def\confYear{2025}
\title{ATAS: Any-to-Any Self-Distillation\\for Enhanced Open-Vocabulary Dense Prediction}

\author{Juan Yeo\thanks{Equal contribution, randomly ordered.} \quad Soonwoo Cha$^*$\thanks{Now at RIST. $^\S$Now at Samsung Research.}\ \quad Jiwoo Song$^*{^\S}$ \quad Hyunbin Jin \quad Taesup Kim\thanks{Corresponding author.}\\
Gradudate School of Data Science, Seoul National University\\
}

\begin{document}
\maketitle
\input{sec/0_abstract}    
\input{sec/1_intro}
\input{sec/3_ATAS}
\input{sec/4_Experiments}
\input{sec/5_related_work}
\input{sec/6_conclusion}
{
    \small
    \bibliographystyle{ieeenat_fullname}
    \bibliography{main}
}




\end{document}

%% file: sec/0_abstract.tex
\begin{abstract}

Vision-language models such as CLIP have recently propelled open-vocabulary dense prediction tasks by enabling recognition of a broad range of visual concepts.
However, CLIP still struggles with fine-grained, region-level understanding, hindering its effectiveness on these dense prediction tasks.
We identify two pivotal factors required to address this limitation: semantic coherence and fine-grained vision-language alignment.
Current adaptation methods often improve fine-grained alignment at the expense of semantic coherence, and often rely on extra modules or supervised fine-tuning.
To overcome these issues, we propose Any-to-Any Self-Distillation (ATAS), a novel approach that simultaneously enhances semantic coherence and fine-grained alignment by leveraging a model’s own knowledge across all representation levels. 
Unlike prior methods, ATAS uses only unlabeled images and an internal self-distillation process to refine CLIP’s representations, preserving local semantic consistency while sharpening local detail recognition. 
On open-vocabulary object detection and semantic segmentation benchmarks, ATAS achieves substantial performance gains, outperforming baseline CLIP models. These results validate the effectiveness of our approach and underscore the importance of jointly maintaining semantic coherence and fine-grained alignment for advanced open-vocabulary dense prediction.

\end{abstract}

%% file: sec/1_intro.tex
\section{Introduction}
\label{sec:intro}
Open-vocabulary prediction in computer vision~\cite{zou2023segment, luddecke2022image, vild, clipself} aims to recognize a wide range of visual concepts beyond a fixed set of categories.
This capability is especially important for dense prediction tasks like object detection and semantic segmentation, where models must handle diverse and previously unseen object classes in complex scenes. 
The advent of contrastive Vision-Language Models (VLMs) such as CLIP~\cite{clip} has significantly advanced this field.
Trained on large-scale image-text pairs, CLIP learns a broad visual semantic space and shows strong zero-shot performance on classification tasks.
However, directly applying CLIP to dense prediction remains challenging because these tasks demand a fine-grained, region-level understanding of images, whereas CLIP’s training is primarily geared towards image-level representation learning.
As a result, CLIP’s performance in segmenting or detecting novel objects is limited by its gap in detailed local feature understanding.


\begin{figure}[t]
    \centering
    \setlength{\tabcolsep}{0.0130\linewidth}
    \begin{subfigure}[b]{1.0\linewidth}
        \centering
        \includegraphics[width=\linewidth]{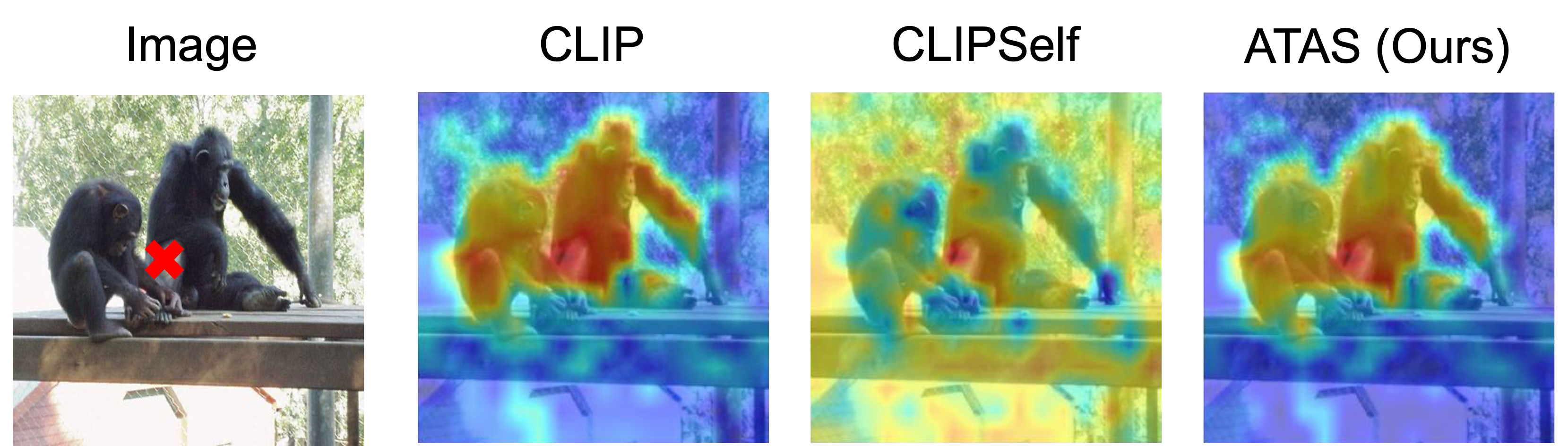}
        \caption{Visualization of semantic coherence via patch similarity}
        \label{fig:intro_subfig1}
        \vspace{3mm}
    \end{subfigure}
    \begin{subfigure}[b]{1.0\linewidth}
        \centering
        \includegraphics[width=\linewidth]{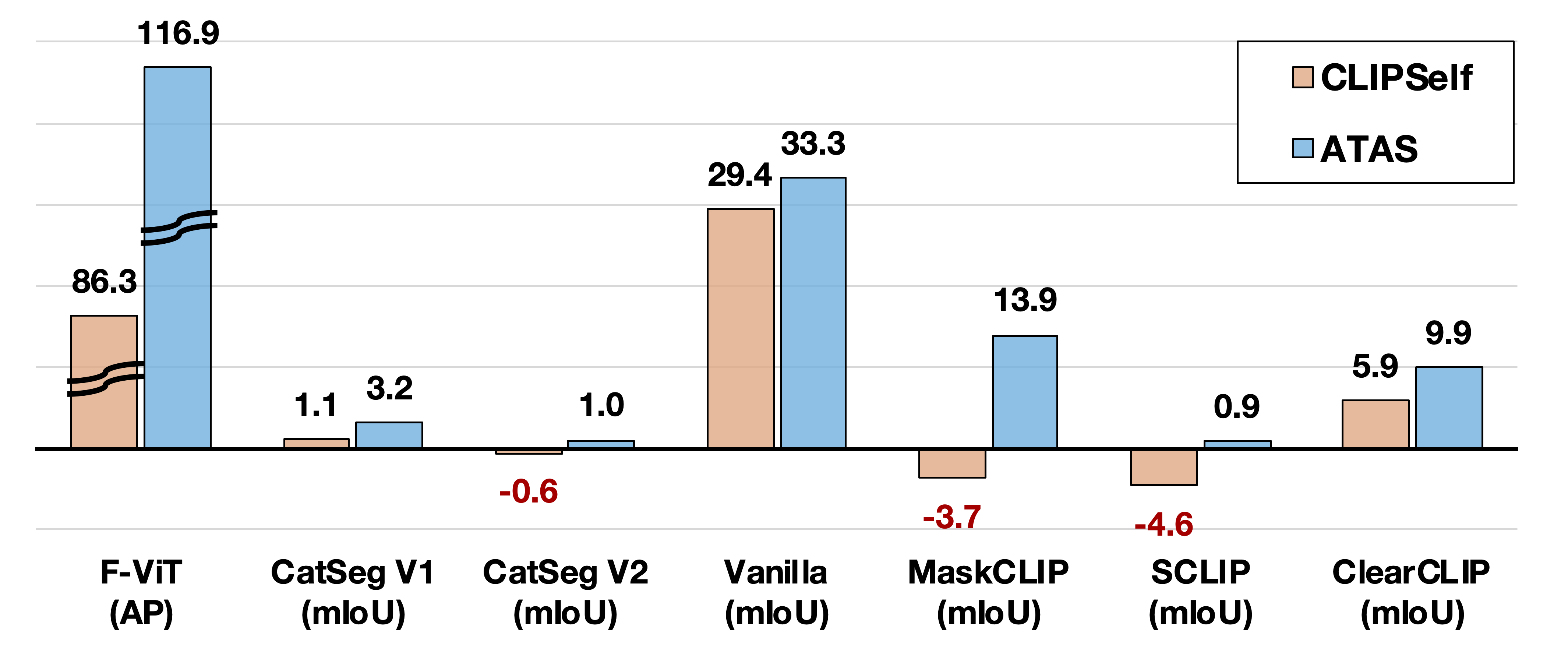}
        \caption{Performance improvement rates achieved by adapting CLIP to dense prediction tasks}
        \label{fig:intro_subfig2}
    \end{subfigure}
    \caption{\textbf{Overall results of our proposed method.} (a) Patch similarity maps illustrate the cosine similarity between a specific patch (marked with a red `$\times$') and other patches, revealing semantic coherence. (b) Downstream dense prediction task performances present relative gains (\%) compared to original CLIP, averaged across datasets.}
    \vspace{-10pt}
    \label{fig:pipeline}
\end{figure}

To address these challenges, recent studies have focused on adapting CLIP for open-vocabulary dense prediction tasks. 
Many of these approaches~\cite{zsseg, zegformer, ovseg, freeseg, maft} involve auxiliary components, such as additional modules or further supervised fine-tuning. 
Conversely, several studies~\cite{clipself, pacl, maskclip, clipsurgery, sclip} have sought to enhance CLIP’s intrinsic capabilities for dense prediction tasks without external supervision.
In particular, two key aspects have emerged as crucial for success: (1) \textbf{semantic coherence} among local image features, and (2) \textbf{fine-grained vision-language alignment} between local regions and text descriptions.
Ideally, a model should maintain consistent semantic features for similar parts of an image (coherence) while also aligning those parts correctly with textual concepts (alignment). 
Existing methods tend to tackle these aspects individually, often improving one at the expense of the other, and thus fall short of fully unlocking CLIP’s potential in dense predictions.


\begin{figure}[t]
  \centering
  \begin{minipage}{0.52\columnwidth}
    \centering
    \includegraphics[width=\linewidth]{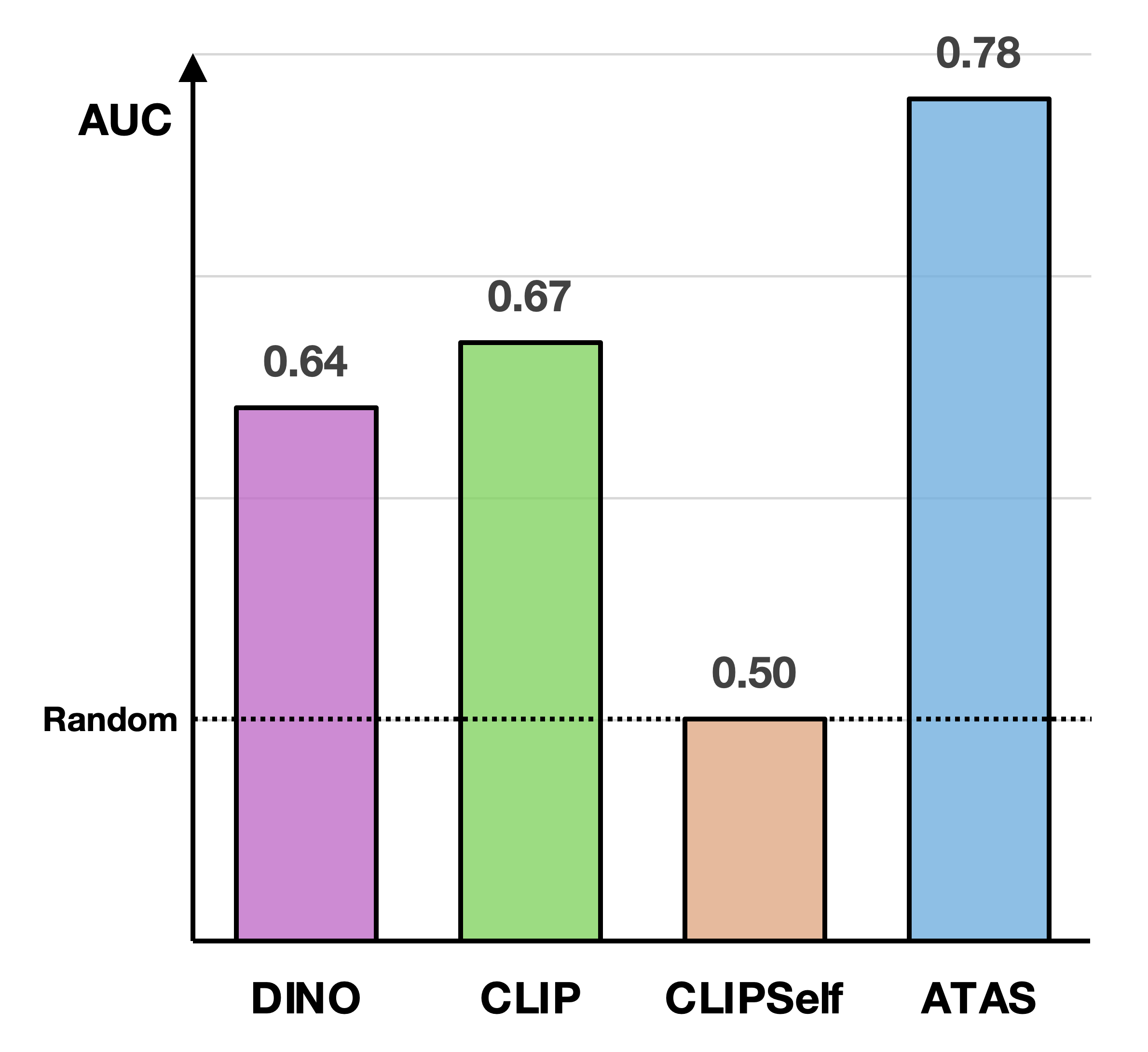}
    \vspace{-1.5em}
    \caption*{(a) Semantic Coherence}
  \end{minipage}
  \hfill
  \begin{minipage}{0.45\columnwidth}
    \centering
    \includegraphics[width=\linewidth]{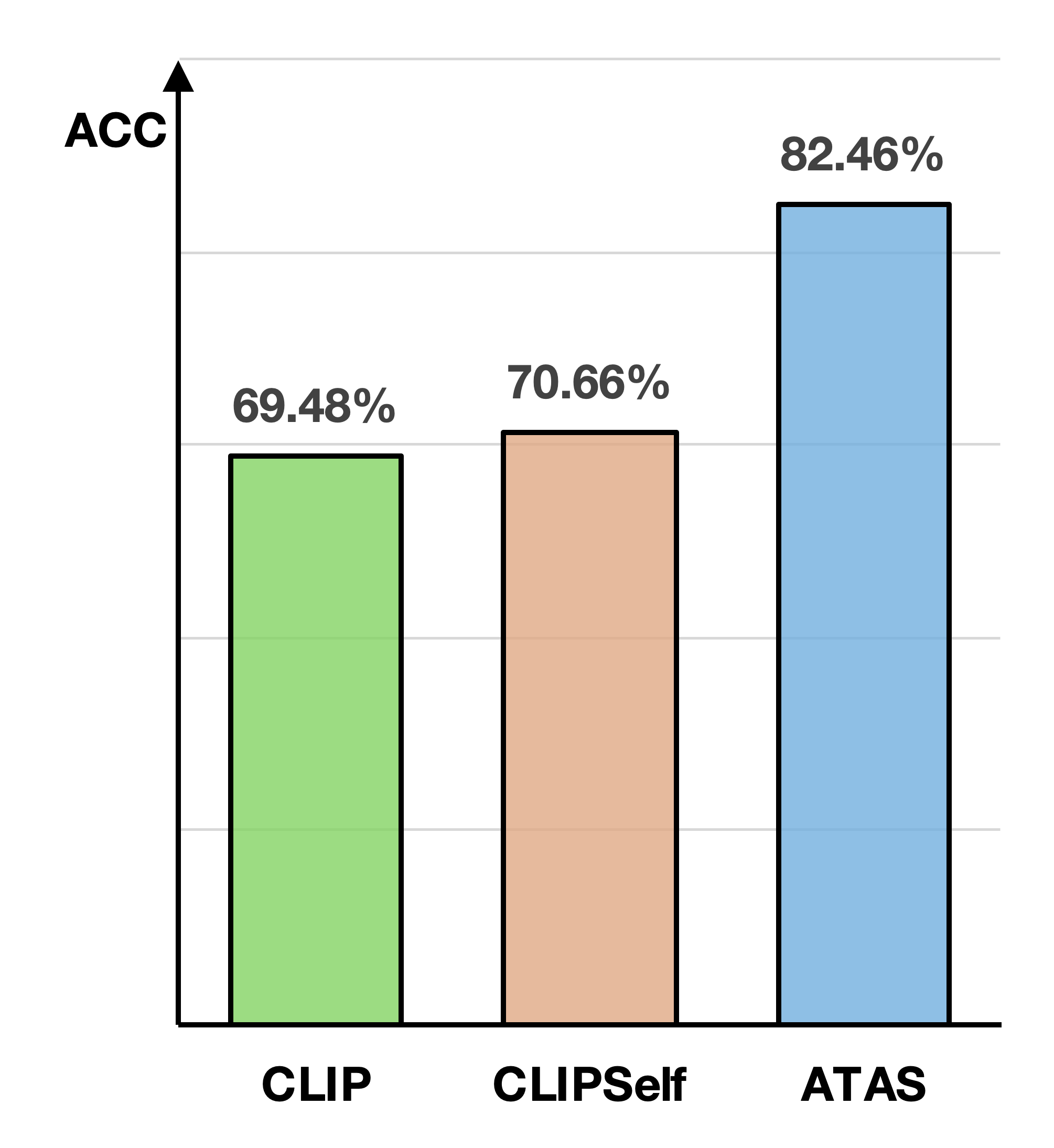}
    \vspace{-1.6em}
    \caption*{(b) Fine-grained Alignment}
  \end{minipage}
  \caption{\textbf{Two key factors in adapting CLIP for dense prediction.} (a) Semantic coherence is quantified using AUROC scores in a patch-level binary classification and (b) fine-grained Alignment is measured by patch-level classification accuracy on Pascal VOC~\cite{everingham2015pascal}}
  \label{fig:two_factors}
  \vspace{-1em}
\end{figure}

\textbf{Semantic coherence} refers to the model’s ability to produce similar feature representations for semantically related local regions (e.g., patches in a Vision Transformer~\cite{vit}). 
This property is vital for dense prediction tasks: if patches belonging to the same object or scene element have consistent features, the model can more reliably group them and recognize the object or region.
Maintaining strong coherence helps capture spatial relationships and yields more accurate segmentation masks or detection bounding boxes.
Notably, we find that CLIP’s vision encoder already preserves a reasonable degree of semantic coherence across its patch embeddings on par with self-supervised vision transformers like DINO~\cite{dino}. 
This intrinsic coherence indicates that CLIP provides a sound foundation for dense prediction tasks, as long as this characteristic is preserved or enhanced during fine-tuning.


\textbf{Fine-grained vision-language alignment}, on the other hand, measures how well localized image features correspond to textual concepts. 
While CLIP excels at aligning whole image embeddings with text, it has been observed to fall short in aligning smaller regions or patches with descriptive text prompts~\cite{regionclip, glip, pacl, clipself}.
Our preliminary analysis confirms this limitation: CLIP struggles with patch-level recognition (e.g., classifying small image regions), highlighting a misalignment between local visual features and textual embeddings (see~\cref{fig:two_factors}).
Since open-vocabulary dense prediction tasks rely heavily on local (region or patch-level) representations to identify novel objects categories, improving this fine-grained alignment is essential for optimal performance. 

Addressing both semantic coherence and fine-grained alignment simultaneously is challenging.
Strategies that heavily emphasize fine-grained alignment can inadvertently disrupt the semantic structure of the visual features.
For instance, CLIPSelf~\cite{clipself} that forces each patch embedding to align closely with the image-level embedding (which is well-aligned with text) through self-distillation succeed in training patches to predict text labels, but we observed that they often collapse the variation among patch features, reducing the coherence of the representation.
In our experiments, this trade-off led to poorer segmentation quality – an indication that sacrificing coherence harms the model’s spatial understanding. 
Moreover, some alignment-focused approaches rely on additional data, such as region-caption pairs or external training signals. 
For example, methods like RegionCLIP~\cite{regionclip} and PACL~\cite{pacl} refine CLIP’s local feature alignment by leveraging additional region-level image-text data or projection modules.
However, these approaches typically require extensive labeled or captioned datasets, which can be resource-intensive. 
These observations underscore the need for a new approach that balances and enhances both coherence and alignment without heavy external supervision.

In this work, we propose \textbf{Any-to-Any Self-Distillation (ATAS)}, a novel framework that enhances CLIP’s dense prediction capabilities by simultaneously improving semantic coherence and fine-grained alignment. 
Unlike prior methods that address these aspects separately, we employ a unified self-distillation strategy to transfer knowledge across different representation levels.
By integrating global-to-local, local-to-local, and global-to-global distillation, ATAS refines feature representations at all scales using only unlabeled images, without additional supervision.
Empirically, ATAS yields substantial performance boosts on multiple open-vocabulary benchmarks, including object segmentation and detection, outperforming not only the original CLIP but also previous adaptation techniques.
The main contributions of our work are summarized as follows:
\begin{itemize}[noitemsep,nosep]
 \item \textbf{Insight into key factors.} We identify (1) semantic coherence and (2) fine-grained vision-language alignment as crucial for adapting CLIP to open-vocabulary dense prediction, demonstrating their impact through qualitative and quantitative analysis.
 \item \textbf{Any-to-any self-distillation framework.} We propose Any-to-Any Self-Distillation (ATAS), the holistic approach to enhance both coherence and alignment simultaneously, using only unlabeled images without external supervision.
 \item \textbf{Improved open-vocabulary performance.} Our approach significantly improves open-vocabulary segmentation and detection, consistently outperforming baseline CLIP across architectures and benchmarks, proving its effectiveness for dense prediction tasks.
\end{itemize}

%% file: sec/3_ATAS.tex
\begin{figure*}[htbp]
    \centering
    \includegraphics[width=0.9\linewidth]{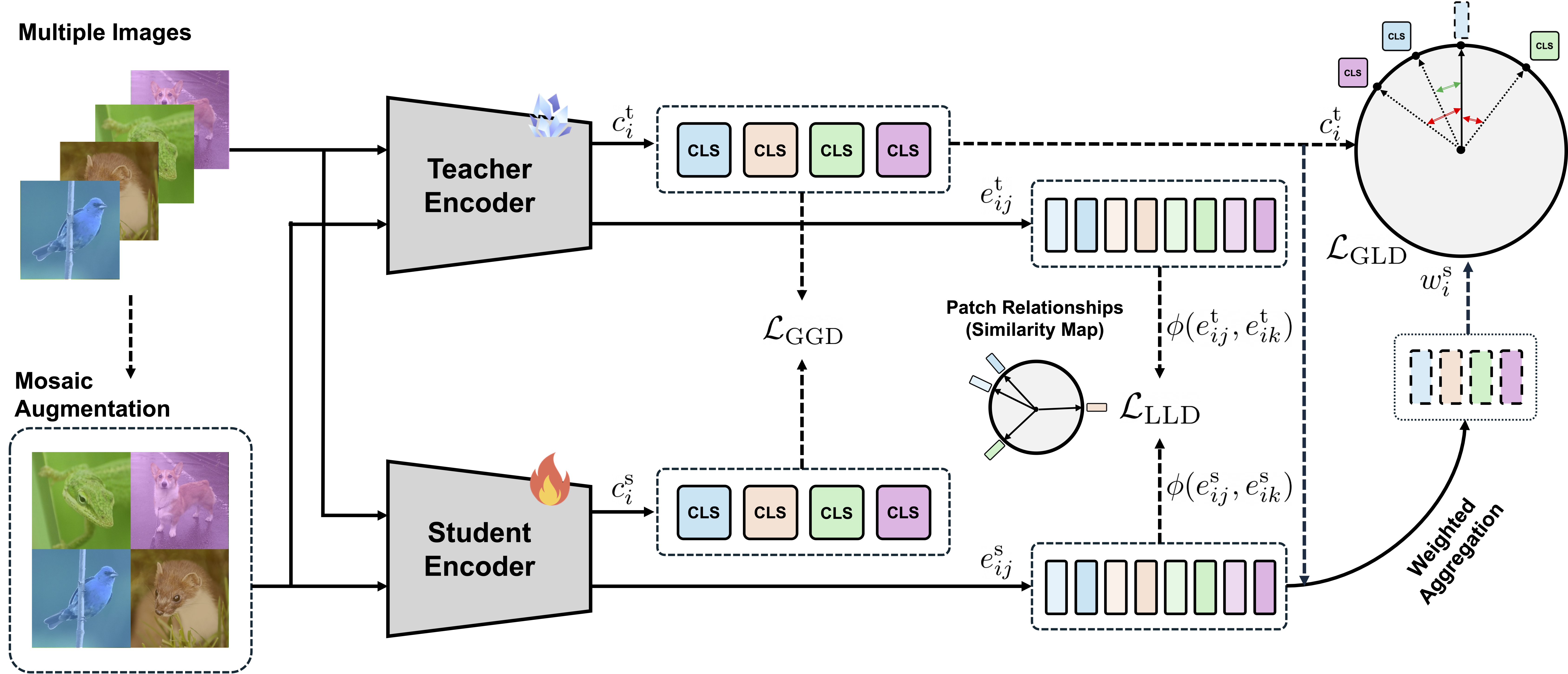}
    \caption{\textbf{Overview of ATAS framework.} ATAS is built upon three core components: (1) Global-to-Local Distillation (2) Global-to-Global Distillation and (3) Local-to-Local Distillation. We utilize patch token embeddings from mosaic-augmented images as local representations and CLS token embeddings from individual images as global representations.}
    \label{fig:method}
\end{figure*}
\section{Any-to-Any Self-Distillation (ATAS)}
The CLIP image encoder $f$, typically based on a Vision Transformer (ViT), represents an image using a set of patch embeddings along with a special class token (CLS) embedding. 
Formally, a given $i^{\text{th}}$ input image  $X_i \in \mathbb{R}^{3\times H\times W}$  is divided into  $n$ non-overlapping patches $\{x_{ij}\}_{j=1}^n$, each of size $P \times P$  pixels.
Each  $j^{\text{th}}$ patch is then mapped to a feature embedding  $e_{ij} \in \mathbb{R}^d$, capturing the local visual representation of that region, while a class token embedding  $c_i \in \mathbb{R}^d$ serves as a global semantic representation of the entire image.
During CLIP’s pre-training, this class token embedding is explicitly optimized to align with text embeddings in a shared vision-language space, facilitating open-vocabulary recognition.

In contrast, patch embeddings  $\{e_{ij}\}_{j=1}^n$ are not directly aligned to text during training—there is no explicit supervision for them.
Instead, they primarily capture local visual features of their respective regions, contributing to the overall image representation through self-attention. 
Our experiments (\cref{fig:two_factors}) show that CLIP’s patch embeddings exhibit high semantic coherence, meaning that patches from the same object or region tend to have similar representations.
This property is beneficial for dense prediction tasks, as it helps in grouping and distinguishing objects within an image.
However, unlike $c_i$, which is explicitly optimized for vision-language alignment, individual patch embeddings lack direct alignment with textual representations, limiting their effectiveness in open-vocabulary dense prediction.

To bridge this gap, we aim to transfer vision-language alignment from the class token $c_i$ to the patch embeddings while preserving their inherent semantic coherence through a self-distillation strategy. 
A naive transfer risks disrupting the natural structure of CLIP’s patch representations, so our goal is to enhance fine-grained textual alignment while maintaining semantic consistency.
To achieve these two complementary goals, we propose Any-to-Any Self-Distillation (ATAS), a novel framework that refines CLIP’s representations without external supervision.
ATAS accomplishes this through a multi-level distillation approach that refines CLIP’s representations without requiring external supervision.
Specifically, ATAS consists of three key distillation mechanisms:
\begin{itemize}[noitemsep,nosep]
    \item \textbf{Global-to-Local Distillation.} Injects vision-language alignment from the class token into patch embeddings, ensuring that local features gain semantic grounding in the shared vision-language space.
    \item \textbf{Local-to-Local Distillation.} Preserves and reinforces the natural semantic coherence among patch embeddings by encouraging the student model to maintain the structural relationships of patch representations.
    \item \textbf{Global-to-Global Distillation.} Ensures that the overall semantic representation of the image remains stable, preventing any degradation in CLIP’s high-level understanding during the adaptation process.
\end{itemize}
By combining these three distillation objectives, ATAS effectively enhances CLIP’s patch representations to be both text-aware and semantically consistent, thereby significantly improving its performance in open-vocabulary dense prediction tasks.
The following sections detail the formulation of each distillation mechanism and its role in refining CLIP’s feature representations.

\subsection{Global-to-Local Distillation}
\label{sec:GLD}
Global-to-Local Distillation (GLD) aims to transfer the class token embedding (teacher), which is aligned with textual representations, into patch embeddings (student) to \textit{enhance fine-grained vision-language alignment}.
A naive approach would be to directly force each patch embedding to match the class token embedding. 
However, this overlooks the fact that not all patches are equally relevant—background regions or unrelated areas should not dominate the alignment. 
Instead, we construct an aggregated local representation that captures the salient content of the image from the student’s patch-level perspective and align it with the teacher’s class token (global) feature.

To achieve this, we perform weighted pooling of the student’s patch embeddings, where the weights reflect how much each patch correlates with the teacher’s global image embedding. 
Formally, for the  $i^{\text{th}}$ image in a batch, let $c_{i}^{\text{t}}$ be the teacher’s global feature (class token embedding from the original CLIP) and  $e_{ij}^{\text{s}}$ be the student’s feature for the  $j^\text{th}$ patch. 
We define the similarity-based local aggregation as follows:

\begin{equation}
    w^{\text{s}}_{i} = \sum_{j=1}^n \text{softmax}\left( 
    \phi(e^{\text{s}}_{ij}, c^{\text{t}}_{i})/\tau
    \right) e^{\text{s}}_{ij},
\end{equation}
where the function $\phi(e, e')$ is the cosine similarity between features $e$ and $e'$. 

We then align the aggregated local feature $w^{\text{s}}_{i}$ with the teacher’s global feature $c_{i}^{\text{c}}$ using a contrastive learning objective. 
Inspired by CLIP’s original image-text alignment, we adapt contrastive loss to our Global-to-Local Distillation setting with a batch of $N$ images:
\begin{equation}
\mathcal{L}_{\mathrm{GLD}} =  \frac{1}{N} \sum_{i=1}^{N}\left[-\log\frac{\text{exp}\left({\phi\left(w^{\text{s}}_{i},c^{\text{t}}_{i}\right)}/\tau\right)}{\sum_{j=1}^{N}\text{exp}({\phi(w^{\text{s}}_{i},c^{\text{t}}_{j})/\tau})}\right].
\end{equation}

To further enhance the alignment between global and local features, we leverage object-centric images during training.
These images, which predominantly contain well-defined objects, help reinforce semantic consistency between class token and patch embeddings.
This approach ensures that salient patches align effectively with the global semantics while minimizing interference from background regions. 
Moreover, we utilize \textit{mosaic augmentation} to ensure multiple objects coexist within a single image and leverage the mosaic pattern to properly guide the alignment between global and local features.
More details on dataset selection and augmentation strategies can be found in~\cref{sec:experiment}.

\subsection{Local-to-Local Distillation}

While Global-to-Local Distillation successfully transfers vision-language alignment from the class token to patch embeddings, this adaptation process introduces unintended feature drift.
In particular, patches that previously maintained a strong semantic relationship may become less consistent, leading to incoherent local representations.
This issue arises because GLD prioritizes fine-grained vision-language alignment, potentially overriding CLIP’s inherent semantic coherence among patches.

To address this, Local-to-Local Distillation (LLD) reinforces semantic consistency across all patches by \textit{distilling CLIP’s pre-existing semantic coherence} into the student model. 
Instead of directly distilling patch embeddings, we focus on preserving the \textit{relational structure}~\cite{rkd, irg} between them.
That is, rather than treating each patch in isolation, we ensure that the relative relationships between patches in the student model remain faithful to those in the teacher model.

Concretely, given a teacher CLIP model, we extract pairwise patch similarities $\phi(e^{\text{t}}_{ij}, e^{\text{t}}_{ik})$ from its patch embeddings and transfer this structural knowledge to the student model by the LLD objective:
\begin{equation}
    \mathcal{L}_{\text{LLD}} = \frac{1}{n^2N} \sum_{i=1}^{N} \sum_{j=1}^{n} \sum_{k=1}^{n} \left(\phi(e^{\text{s}}_{ij}, e^{\text{s}}_{ik}) - \phi(e^{\text{t}}_{ij}, e^{\text{t}}_{ik})\right)^2,
\end{equation}
This not only preserves CLIP’s intrinsic grouping of semantic regions but also complements GLD by preventing alignment-induced feature inconsistencies.

\subsection{Global-to-Global Distillation}
\label{sec:GGD}
In Transformer-based architectures, the class token plays a critical role in aggregating and representing global information. 
In CLIP, this token is particularly significant as it serves as the primary vision-language anchor, encoding high-level semantics that align with text representations.
Given its central role, any modifications to patch-level features, such as those introduced by Global-to-Local Distillation (GLD), could inadvertently disrupt the class token’s representation, leading to a loss of semantic coherence and weakened alignment with textual concepts.

To mitigate this, we introduce Global-to-Global Distillation (GGD), ensuring that the student model retains the semantic fidelity of the teacher’s class token embedding. 
This global-to-global distillation loss function is defined as follows:
\begin{equation}
    \mathcal{L}_{\mathrm{GGD}} =  \frac{1}{N} \sum_{i=1}^{N}\left[-\log\frac{\text{exp}{(\phi(c^{\text{s}}_{i}, c^{\text{t}}_{i})/\tau)}}{\sum_{j=1}^{N}\text{exp}{(\phi(c^{\text{s}}_{i}, c^{\text{t}}_{j})/\tau)}}\right],
\end{equation}
where we apply contrastive objective as in original CLIP.
Our experiments confirm that explicitly distilling this global representation leads to performance improvements in dense prediction tasks, suggesting that preserving the class token’s role benefits both fine-grained alignment and semantic structure maintenance.

To sum up, the overall objective function for training the student model is formulated by combining the aforementioned distillation losses as follows:
\begin{equation}
\mathcal{L} = 
\lambda_{1}\cdot\mathcal{L}_{\mathrm{GLD}} + \lambda_{2}\cdot\mathcal{L}_{\mathrm{LLD}}  + \lambda_{3}\cdot\mathcal{L}_{\mathrm{GGD}}.
\end{equation}
where \(\lambda_{1}\) \(\lambda_{2}\) and \(\lambda_{3}\) are the hyper-parameters used to balance between losses.

%% file: sec/4_Experiments.tex
\input{tables/ovss_notraining}

\section{Experiments}
\label{sec:experiment}

\subsection{Experimental Setup}
\paragraph{Datasets}
We applied ATAS to the CLIP image encoder using the ImageNet~\cite{imagenet} dataset, which comprises 1.2 million object-centric training images.
To evaluate performance, we conducted two open-vocabulary dense prediction tasks: semantic segmentation and object detection. For semantic segmentation, we tested two scenarios: (1) a zero-shot scenario with no additional training and (2) a fine-tuning scenario where CLIP was further trained on the COCO-Stuff~\cite{cocostuff} dataset with segmentation labels. Both settings were evaluated on ADE20K~\cite{ade20k}, Pascal VOC~\cite{everingham2015pascal}, and Pascal Context~\cite{pascalcontext}, with COCO-Stuff and Cityscapes~\cite{cityscapes} additionally used in the zero-shot setting.
For object detection, we employed the OV-COCO benchmark, which transforms the COCO~\cite{coco} into an open-vocabulary setting by designating 48 base classes and 17 novel classes, and the OV-LVIS benchmark, which considers the 337 rare categories in LVIS v1.0~\cite{lvis} as novel categories, as proposed in ViLD~\cite{gu2021open}.

\paragraph{Implementation Details} \label{experiments:implementation_details}
We used AdamW~\cite{adamw} optimizer with a learning rate of 
$1\mathrm{e}{-5}$, a weight decay of 0.1, and 4 workers. We trained for 6 epochs using 4 RTX 3090 GPUs, with a batch size of 36 per GPU. In this paper, we trained CLIP model~\cite{clip} using an ATAS method and then applied it to the methodologies used for each task. During training, we trained only image encoder. We set hyperparameters \(\lambda_1=1\), \(\lambda_2=0.01\), \(\lambda_3=1\), \(\tau=1\) for overall experiments.

\paragraph{Mosaic Augmentation}
Effective knowledge transfer from the teacher class token to the student patch tokens demands that the class tokens remain robust, even in scenes containing multiple objects. Cropping scene-centric images helps isolate single-object semantics within multi-object scenes but may also weaken the class token by capturing only partial objects, resulting in ambiguous representations. To mitigate this limitation, we adopt mosaic augmentation, where multiple object-centric images are concatenated to form high-resolution composites. Specifically, we obtain the teacher CLS tokens from individual single-object images while extracting patch tokens from the mosaic images, which better reflect the complexity of real-world scenes. By separating the source of the class token (single-object) from the source of the patch tokens (multi-object), our method retains clear global semantics while training robust local feature representations for dense prediction. Further details are provided in the Appendix.


\input{tables/ovss_catseg_main}

\subsection{Open-Vocabulary Semantic Segmentation} \label{experiment:OVSS}
\subsubsection{Zero-shot Scenarios}
\paragraph{Baselines and Settings}
To directly evaluate CLIP's dense prediction capabilities without introducing additional modules or supervised training, we performed zero-shot semantic segmentation using the OpenAI CLIP ViT-B model. Our experiments covered four approaches: vanilla CLIP~\cite{clip}, MaskCLIP~\cite{maskclip}, SCLIP~\cite{sclip}, and ClearCLIP~\cite{clearclip}. Unlike vanilla CLIP, the other variants make minor modifications to the final layer of the CLIP vision transformer. MaskCLIP extracts value embeddings from the last self-attention block, while SCLIP enhances self-attention by leveraging query-query and key-key correlations. ClearCLIP removes the last layer's residual connection during dense feature extraction. Since our method was trained with a residual connection in the last layer, we evaluated ClearCLIP with a modified version with the connection, denoted as ClearCLIP\(^{\dag}\).

\paragraph{Experimental Results}
In our zero-shot semantic segmentation experiments, ATAS consistently improved the performance of the CLIP encoder, as summarized in \cref{tab:notraining}. Specifically, ATAS increased the average mIoU by 4.2 in the Vanilla setting and by 3.8 in the MaskCLIP compared to the original CLIP. Notably, while CLIPSelf occasionally exhibits performance degradation after training, our method reliably enhances CLIP's dense prediction capabilities. Furthermore, the qualitative results in \cref{fig:zero_visualization_main} demonstrate that our approach successfully segment objects that CLIP frequently fails to capture, underscoring the robust and consistent improvements offered by ATAS.

\input{tables/figure_zero_shot}
\vspace{0.3em}
\subsubsection{Fine-tuning Scenarios}
\label{sec:finetuning-ss}
\paragraph{Baselines and Settings}
In this experiment, we used two versions of the CAT-Seg model: CAT-Seg v1 and CAT-Seg v2. CAT-Seg v1 is used in CLIPSelf~\cite{clipself} to evaluate the performance of open-vocabulary semantic segmentation. CAT-Seg v2~\cite{catseg} is a revised version of CAT-Seg v1.
The detailed differences between the two models are explained in the Appendix.
Both models leverage the dense features of CLIP ViTs within their cost-aggregation framework and are fully fine-tuned on the COCO Stuff~\cite{cocostuff} dataset.
To ensure a fair performance comparison, we adopted CAT-Seg’s default settings. In this experiment, we replaced the CLIP weights in both versions of CAT-Seg with three different models: CLIP, CLIPSelf, and ATAS. Each model was then trained using the CAT-Seg approach. The experiments were conducted using OpenAI CLIP ViT-B and ViT-L, with the performance results for ViT-L provided in the Appendix.

\paragraph{Experimental Results}
 Our results, summarized in \cref{tab:catseg_merged}, demonstrate that ATAS consistently enhances the performance of CAT-Seg across multiple datasets and versions. Notably, ATAS achieves improvements in all scenarios except one, including a 3.6 mIoU increase on the ADE-150 dataset. By contrast, the performance of CLIPSelf varies significantly, particularly under CAT-Seg v2. While CLIPSelf achieves marginal improvements with CAT-Seg v1 compared to the original CLIP, it unexpectedly experiences performance degradation with CAT-Seg v2. This variability indicates that CLIPSelf is sensitive to specific training setups, whereas ATAS maintains stable performance gains regardless of the CAT-Seg variant employed. \\
 The magnitude of improvements observed (0.5-1.5 mIoU) for both ATAS and CLIPSelf is modest. The primary reason for this is CAT-Seg's extensive supervised fine-tuning using detailed segmentation annotations, which already extracts much of the potential performance benefits from the underlying CLIP model. Despite these constraints, the consistent performance improvement achieved by ATAS underscores its robustness and effectiveness in enhancing CLIP backbone.

\input{tables/ovod_eva_full_main}

\begin{table*}[ht]
\centering
\vspace{0.3em}
\begin{minipage}[t]{0.305\linewidth}
    \begin{subtable}{\linewidth}
        \resizebox{\linewidth}{!}{%
        \input{tables/ablation_5_alignment}}
        \caption{Distillation Losses}
        \label{table:ablation_losses}
    \end{subtable}
\end{minipage}
\hfill
\begin{minipage}[t]{0.3\linewidth}
    \begin{subtable}{\linewidth}
        \resizebox{\linewidth}{!}{%
        \input{tables/ablation_3_weight_contrastive}}
        \caption{Aligning Methods in GLD}
        \label{table:ablation_weight_contrastive}
    \end{subtable}
\end{minipage}
\hfill
\begin{minipage}[t]{0.36\linewidth}
    \begin{subtable}{\linewidth}
        \resizebox{\linewidth}{!}{%
        \input{tables/ablation_2_hyperparam}}
        \caption{Hyperparameter Scale}
        \label{table:ablation_hyperparam}
    \end{subtable}
\end{minipage}
\caption{\textbf{Ablation Studies of ATAS.}}
\end{table*}

\subsection{Open-Vocabulary Object Detection}
\paragraph{Baselines and Settings}
We employed the F-ViT architecture from CLIPSelf~\cite{clipself}, a two-stage detector that integrates a Feature Pyramid Network (FPN) with a frozen CLIP backbone as introduced in F-VLM~\cite{fvlm}. In our experiments, we replaced the original CLIP weights with those trained using ATAS.

In CLIPSelf, the EVA-CLIP model~\cite{sun2023eva} was selected for open-vocabulary object detection (OVOD) due to its superior efficiency and capacity. Following this approach, we conducted our OVOD experiments using the EVA-CLIP ViT-B model, training F-ViT for 2 epochs on OV-COCO and for 36 epochs on OV-LVIS with a batch size of 64, which yielded the best performance.

\paragraph{Experimental Results}
For evaluation, we used the box AP at an IoU threshold of 0.5 for novel categories (AP\(^\text{novel}_{50}\)), as in previous studies on the OV-COCO dataset, and the mean mask AP for rare categories (mAP\(_\text{r}\)) on the OV-LVIS dataset. As shown in \cref{tab:ovod_eva_full_main}, ATAS significantly improves detection performance of the baseline CLIP encoder by 19.7 $\text{AP}^{\text{novel}}_{50}$ on OV-COCO and 14.3 $\text{mAP}_{r}$ on OV-LVIS. It outperforms the previous method by 3.6 AP\(^\text{novel}_{50}\) and demonstrated even higher performance compared to other methods on OV-COCO. Additionally, our method achieved a 0.5 mAP\(_\text{r}\) improvement over previous approaches on the OV-LVIS benchmark. However, it exhibits somewhat lower performance when compared to a method that leverages a large-scale training dataset.

\subsection{Ablation Study}
We conducted ablation studies on MaskCLIP and F-ViT using OpenAI ViT-B under the same settings as the above experiments. Our ablation study in ATAS focused on three critical aspects: (a) how different distillation losses influence performance (b) the contribution of contrastive objectives and selective weighting in GLD (c) the impact of various mosaic augmentation strategies to find the best setup. The detailed experimental settings are provided in the Appendix.
\vspace{-0.5em}

\paragraph{Effectiveness of Distillation Losses}
In this paper, we propose three distillation losses: $\mathcal{L}_{\mathrm{GLD}}$, $\mathcal{L}_{\mathrm{LLD}}$ and $\mathcal{L}_{\mathrm{GGD}}$. We conducted an ablation study to analyze the contribution of each distillation component to the downstream tasks. The results are presented in \cref{table:ablation_losses}.\\
When employing only the baseline CLIP model (without any distillation), the performance is notably limited, achieving 34.28 mIoU in segmentation and 31.40 $\text{AP}_{50}$ in detection. Introducing Global-to-Local Distillation ($\mathcal{L}_{\mathrm{GLD}}$) significantly improves both segmentation (+3.37) and detection (+14.58), confirming its effectiveness in transferring CLIP's global vision-language alignment into patch embeddings. Adding Local-to-Local Distillation ($\mathcal{L}_{\mathrm{LLD}}$) further enhances segmentation performance slightly but shows a marginal degradation in detection, suggesting that preserving local semantic consistency alone may constrain certain discriminative aspects of patch representations. Crucially, the final addition of Global-to-Global Distillation ($\mathcal{L}_{\mathrm{GGD}}$) reconciles these trade-offs, providing the model with a balanced improvement across both segmentation (+5.06) and detection (+14.98) tasks by ensuring the overall global semantic integrity remains intact.
\vspace{-0.5em}
\paragraph{Effectiveness of Contrastive Distillation}
The ablation results presented in \cref{table:ablation_weight_contrastive} dissect the role of two essential factors in Global-to-Local Distillation ($\mathcal{L}_{\mathrm{GLD}}$): (1) the contrastive objective, and (2) the selective weighting of patches. Initially, employing a cosine-similarity objective (weighted or not) underperforms compared to contrastive-based methods, highlighting contrastive objective's advantage in capturing richer semantic distinctions. Moreover, introducing selective weighting further improves both segmentation and detection tasks, implying that the weighting strategy effectively filters out less relevant patches and enables effective learning. The combined use of contrastive loss and selective weighting thus emerges as the optimal solution, providing precise vision-language alignment at the patch level without sacrificing overall semantic coherence.


\paragraph{Effectiveness of Hyperparameters}

We performed ablation studies to validate our hyperparameters, revealing their distinct contributions.
For the loss weights, the LLD weight, $\lambda_{LLD}$, exhibits a clear trade-off: smaller values benefit detection by emphasizing holistic features, while larger values enhance segmentation by preserving fine-grained representations. We fix $\lambda_{LLD}$ to $1e-2$ to maintain a versatile, unified backbone. The GGD loss, $\lambda_{GGD}$, is found to be critical, as its removal causes the student's ImageNet accuracy from 67.25\% to 40.92\%, demonstrating its essential role in grounding the student with foundational knowledge leading performance gains shown in \cref{table:ablation_hyperparam}. 
For the contrastive temperature $\tau$, overall performance is optimized at $\tau=1$, with any deviation leading to a performance degradation.

%% file: tables/ovss_notraining.tex
\begin{table*}[t]
\centering
\resizebox{0.9\linewidth}{!}{ 
\begin{tabular}{cccccccccccccc}
\toprule
\multirow{2}{*}{Method} & 
\multirow{2}{*}{Model}  & 
\multicolumn{2}{c}{VOC20} & 
\multicolumn{2}{c}{PC-59} & 
\multicolumn{2}{c}{COCO-Stuff} & 
\multicolumn{2}{c}{ADE20} & 
\multicolumn{2}{c}{CityScapes} & 
\multicolumn{2}{c}{{Average}} \\
\cmidrule(lr){3-4}
\cmidrule(lr){5-6}
\cmidrule(lr){7-8}
\cmidrule(lr){9-10}
\cmidrule(lr){11-12}
\cmidrule(lr){13-14}
& &
mIoU&mAcc&
mIoU&mAcc&
mIoU&mAcc&
mIoU&mAcc&
mIoU&mAcc&
mIoU&mAcc\\
\midrule\midrule
\multirow{3}{*}{Vanilla~\cite{clip}}  
&CLIP& 41.8   & 57.2   & 9.2   & 21.8   & 4.4    & 11.6     & 2.1   & 7.6   & 5.5   & 12.8    & 12.6  & 22.2   \\
& CLIPSelf & {\ul 53.7} & {\ul 71.9} & {\ul 10.9} & {\ul 28.8} & {\ul 5.5}  & {\ul 18.2}  & \textbf{3.5} & {\ul 13.2} & \textbf{7.7} & {\ul 15.4} & {\ul 16.3} & {\ul 29.5} \\
& ATAS & \textbf{56.0} & \textbf{75.9} & \textbf{11.9} & \textbf{31.2} & \textbf{6.4} & \textbf{20.6} & {\ul 3.4} & \textbf{15.2} & {\ul 6.6} & \textbf{16.1} & \textbf{16.8} & \textbf{31.8} \\
\midrule
\multirow{3}{*}{MaskCLIP~\cite{maskclip}}
& CLIP& 60.2  & 75.6   & {\ul 26.2} & 47.1 & {\ul 16.5} & 33.4 & 12.2  & 31.3   & {\ul 21.9} & {\ul 49.7} & {\ul 27.4} & 47.4 \\
& CLIPSelf & {\ul 60.7} & {\ul 77.3} & 24.9  & {\ul 50.4}   & 15.8   & {\ul 37.4}    & {\ul 12.9} & {\ul 35.8} & 17.8   & 41.6   & 26.4  & {\ul 48.5}   \\
& ATAS& \textbf{70.2} & \textbf{85.6} & \textbf{29.1} & \textbf{57.4} & \textbf{18.8} & \textbf{43.5} & \textbf{14.7} & \textbf{41.6} & \textbf{23.2} & \textbf{52.7} & \textbf{31.2} & \textbf{56.1} \\ 
\midrule
\multirow{3}{*}{SCLIP~\cite{sclip}}
& CLIP& 78.2 & 89.7 & \textbf{33.0} & 52.2 & {\ul 21.1} & 37.3 & 14.6  & 30.7   & \textbf{29.1} & {\ul 48.1} & {\ul 35.2} & 51.6 \\
& CLIPSelf & 78.2  & {\ul 89.9}   & 31.1  & {\ul 56.5}   & 19.1    & {\ul 44.6} & \textbf{15.4} & \textbf{37.4} & 24.0 & 42.8   & 33.6  & {\ul 54.2}   \\
& ATAS& \textbf{80.6} & \textbf{92.6} & \textbf{33.0} & \textbf{58.0} & \textbf{21.4} & \textbf{45.3} & {\ul 15.0} & \textbf{37.4} & {\ul 27.5} & \textbf{48.6} & \textbf{35.5} & \textbf{56.4} \\
\midrule
\multirow{3}{*}{ClearCLIP\(^{\dag}\)~\cite{clearclip}} 
& CLIP & 68.4  & 81.6   & 24.9  & 40.9   & 14.7   & 26.2  & 7.6   & 18.3 & \textbf{20.8}  & {\ul 33.3}  & 27.3  & 40.1   \\
& CLIPSelf & {\ul 74.2} & {\ul 87.5} & {\ul 25.7} & {\ul 49.9} & {\ul 15.5} & {\ul 38.0} & \textbf{10.9} & {\ul 29.2} & 18.1  & \textbf{34.1}   & {\ul 28.9} & {\ul 47.7} \\
& ATAS & \textbf{75.2} & \textbf{90.3} & \textbf{27.1} & \textbf{51.6} & \textbf{17.3} & \textbf{39.4} & {\ul 10.2} & \textbf{37.0} & {\ul 20.1} & 29.9 & \textbf{30.0} & \textbf{49.6} \\ 
\bottomrule
\end{tabular}
}
\caption{\textbf{Results of open-vocabulary semantic segmentation in \textit{zero-shot} scenarios.} The model that achieved the best performance is marked in \textbf{bold}, while the model with the second highest performance is marked with {\ul underline}}
\label{tab:notraining}
\end{table*}

%% file: tables/ovss_catseg_main.tex

\begin{table*}[t]
\centering
\resizebox{0.85\textwidth}{!}{
\begin{tabular}{l|cccccc|c}
\Xhline{1.0\arrayrulewidth}
\hline
Method & ADE-150 & ADE-847 & PC-59 & PC-459 & PAS-20 & PAS-20b & Average \\
\hline
\hline
SAN~\cite{san}  & 27.5 & 10.1 & 53.8 & 12.6 & 94.0 & -    & -    \\
SED~\cite{sed}  & 31.6 & 11.4 & 57.3 & 18.6 & 94.4 & -    & -    \\
\hline
CAT-Seg (v1)          & 27.2 & 9.0 & 57.5 & 16.7 & 93.7 & 78.3 & 47.1 \\
CAT-Seg (v1) + CLIPSelf & 30.1 & 9.6 & 57.9 & 17.3 & 93.2 & 77.3 & 47.6 \\
CAT-Seg (v1) + ATAS     & 30.8 & 10.4 & \textbf{58.9} & 18.1 & 94.2 & \textbf{78.9} & 48.6 \\
\hline
CAT-Seg (v2)      & 31.6 & 11.9 & 57.4 & 18.9 & 94.7 & 77.0 & 48.6 \\
CAT-Seg (v2) + CLIPSelf & 31.7 & 12.4 & 56.7 & 18.6 & 94.5 & 75.8 & 48.3 \\
CAT-Seg (v2) + ATAS     & \textbf{32.4} & \textbf{12.7} & 58.0 & \textbf{19.6} & \textbf{95.1} & 76.8 & \textbf{49.1} \\
\Xhline{1.0\arrayrulewidth}
\end{tabular}
}
\caption{\textbf{Results of open-vocabulary semantic segmentation (mIoU) in a \textit{fine-tuning} scenario with CAT-Seg.}}
\label{tab:catseg_merged}
\end{table*}

%% file: tables/figure_zero_shot.tex
\begin{figure}[!t]
    \centering
    \begin{subfigure}{0.23\linewidth}
        \centering
        \includegraphics[width=\linewidth]{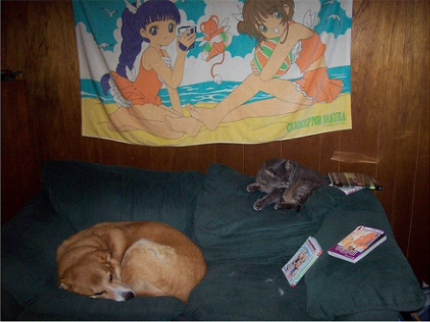}
        \vspace{-4mm}
    \end{subfigure}
    \begin{subfigure}{0.23\linewidth}
        \centering
        \includegraphics[width=\linewidth]{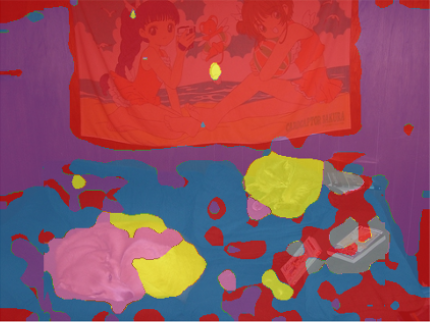}
        \vspace{-4mm}
    \end{subfigure}
    \begin{subfigure}{0.23\linewidth}
        \centering
        \includegraphics[width=\linewidth]{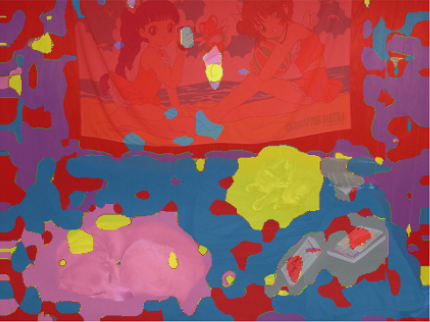}
        \vspace{-4mm}
    \end{subfigure}
    \begin{subfigure}{0.23\linewidth}
        \centering
        \includegraphics[width=\linewidth]{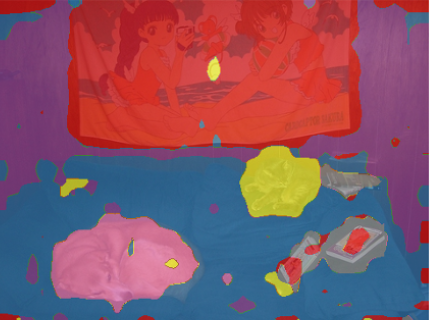}
        \vspace{-4mm}
    \end{subfigure}
    \vspace{0.8mm}
    \begin{subfigure}{0.23\linewidth}
        \centering
        \includegraphics[width=\linewidth]{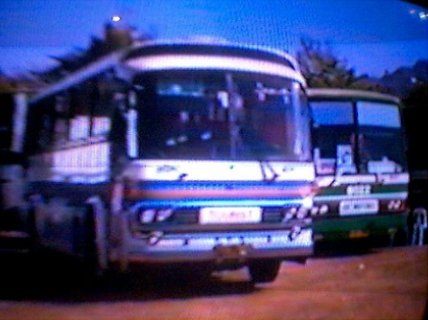}
        \vspace{-5mm}
    \end{subfigure}
    \begin{subfigure}{0.23\linewidth}
        \centering
        \includegraphics[width=\linewidth]{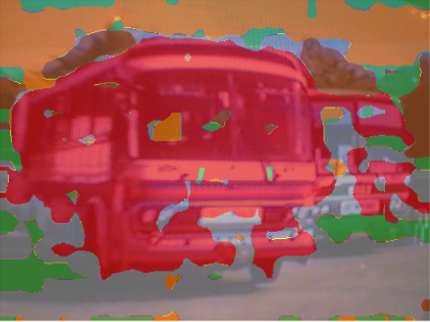}
        \vspace{-5mm}
    \end{subfigure}
    \begin{subfigure}{0.23\linewidth}
        \centering
        \includegraphics[width=\linewidth]{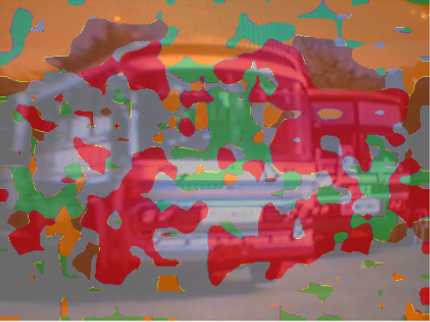}
        \vspace{-5mm}
    \end{subfigure}
    \begin{subfigure}{0.23\linewidth}
        \centering
        \includegraphics[width=\linewidth]{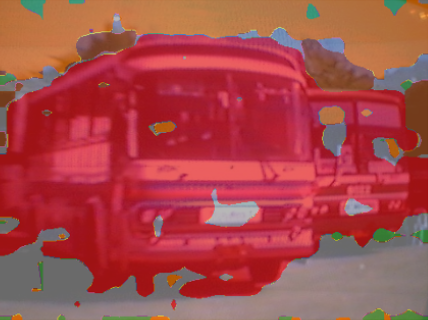}
        \vspace{-5mm}
    \end{subfigure}
    \vspace{1mm}
    \begin{subfigure}{0.23\linewidth}
        \centering
        \includegraphics[width=\linewidth]{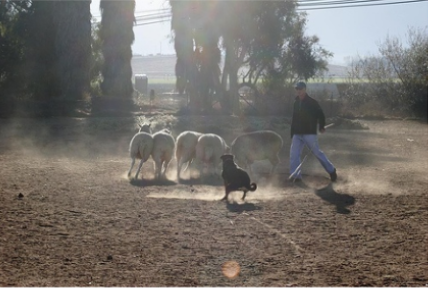}
        \caption{Image}
        \vspace{-3mm}
    \end{subfigure}
    \begin{subfigure}{0.23\linewidth}
        \centering
        \includegraphics[width=\linewidth]{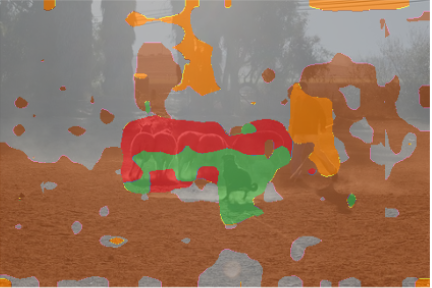}
        \caption{CLIP}
        \vspace{-3mm}
    \end{subfigure}
    \begin{subfigure}{0.23\linewidth}
        \centering
        \includegraphics[width=\linewidth]{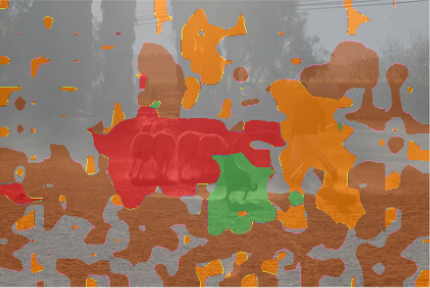}
        \caption{CLIPSelf}
        \vspace{-3mm}
    \end{subfigure}
    \begin{subfigure}{0.23\linewidth}
        \centering
        \includegraphics[width=\linewidth]{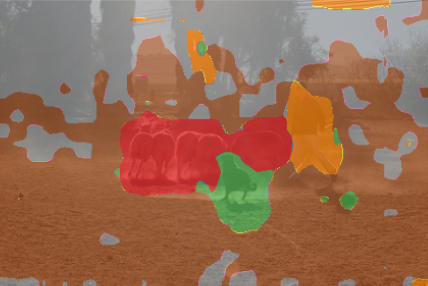}
        \caption{ATAS}
        \vspace{-3mm}
    \end{subfigure}

    \caption{\textbf{Qualitative results on zero-shot semantic segmentation.} We use images from PASCAL VOC and show their segmentations for CLIP, CLIPSelf and ATAS.}
    \vspace{-4mm}
    \label{fig:zero_visualization_main}
\end{figure}

%% file: tables/ovod_eva_full_main.tex
\begin{table*}[t]
\centering
\resizebox{0.7\linewidth}{!}{
\begin{tabular}{l|ccc|cccc} 
\Xhline{1.0\arrayrulewidth}
\hline
    \multirow{2}{*}{Method} & \multicolumn{3}{c|}{OV-COCO} & \multicolumn{4}{c}{OV-LVIS} \\
    & \textcolor{gray}{$\text{AP}_{50}$} & \textcolor{gray}{$\text{AP}^{\text{base}}_{50}$} & $\text{AP}^{\text{novel}}_{50}$ & \textcolor{gray}{$\text{mAP}$} & \textcolor{gray}{$\text{mAP}_{c}$} & \textcolor{gray}{$\text{mAP}_{f}$} & $\text{mAP}_{r}$ \\
    \hline\hline
    CLIM~\cite{clim} & \textcolor{gray}{-} & \textcolor{gray}{42.5} & 25.7 & \textcolor{gray}{-} & \textcolor{gray}{25.6} & \textcolor{gray}{26.7} & 20.8 \\
    ViLD~\cite{vild}       &   \textcolor{gray}{51.3} &   \textcolor{gray}{59.5} &   27.6  &   \textcolor{gray}{25.5} &   \textcolor{gray}{24.6}    &   \textcolor{gray}{30.3} &	16.6 \\
    Detic~\cite{zhou2022detecting}       &   \textcolor{gray}{45.0} &   \textcolor{gray}{47.1} &   27.8 &   \textcolor{gray}{32.4} &   \textcolor{gray}{-}   &   \textcolor{gray}{-} &   24.9 \\
    VLDet~\cite{lin2022learning}     &   \textcolor{gray}{45.8} &   \textcolor{gray}{50.6} &   32.0   &   \textcolor{gray}{30.1} &   \textcolor{gray}{29.8}    &   \textcolor{gray}{32.3}  &	21.7 \\
    DenseVLM~\cite{densevlm} & \textcolor{gray}{47.4} & \textcolor{gray}{52.5} & 33.1 & \textcolor{gray}{21.4} & \textcolor{gray}{18.4} & \textcolor{gray}{-} & 23.9 \\
    BARON-KD~\cite{wu2023baron}  & \textcolor{gray}{53.5} & \textcolor{gray}{60.4} & 34.0 & \textcolor{gray}{27.6} & \textcolor{gray}{27.6} & \textcolor{gray}{29.8} & 22.6 \\
    OADP~\cite{oadp}     &   \textcolor{gray}{50.5} &	\textcolor{gray}{55.8} &   \uline{35.6}  &   \textcolor{gray}{26.6} & \textcolor{gray}{26.3}    &   \textcolor{gray}{29.0} &	21.7 \\
    RO-ViT~\cite{kim2023region}   &   \textcolor{gray}{41.5} &   \textcolor{gray}{-} &  30.2  &   \textcolor{gray}{30.2}  &   \textcolor{gray}{-}   &   \textcolor{gray}{-} & \textbf{28.0} \\
    PromptOVD~\cite{song2023prompt} & \textcolor{gray}{54.9} & \textcolor{gray}{63.5} & 30.6 & \textcolor{gray}{24.2} & \textcolor{gray}{-} & \textcolor{gray}{-}   & 23.1 \\
    
    
    \hline
    F-ViT~\cite{clipself}  & \textcolor{gray}{34.9} & \textcolor{gray}{41.0} & 17.5 & \textcolor{gray}{15.4} & \textcolor{gray}{12.3} & \textcolor{gray}{20.6}  & 11.5 \\
    F-ViT + CLIPSelf~\cite{clipself} & \textcolor{gray}{48.8} & \textcolor{gray}{54.2} & 33.6 & \textcolor{gray}{25.2} & \textcolor{gray}{21.8} & \textcolor{gray}{29.1} & 25.3 \\
    F-ViT + ATAS  & \textcolor{gray}{50.2} & \textcolor{gray}{54.8} & \textbf{37.2} & \textcolor{gray}{25.1} & \textcolor{gray}{21.4} & \textcolor{gray}{28.9} & \uline{25.8} \\
    \Xhline{1.0\arrayrulewidth}
    \hline
    \end{tabular}
}
\caption{\textbf{Results of open-vocabulary object detection on OV-COCO ($\text{AP}^{\text{novel}}_{50}$) and OV-LVIS ($\text{mAP}_{r}$).} The open-vocabulary setting applies only to $\text{AP}^{\text{novel}}_{50}$ for OV-COCO and $\text{mAP}_{r}$ for OV-LVIS, while other metrics are not included in this setting.}
\label{tab:ovod_eva_full_main}
\end{table*}

%% file: tables/ablation_5_alignment.tex
\begin{minipage}{\linewidth}
\centering
\resizebox{\textwidth}{!}{%
\begin{tabular}{ccc|c|c}
\Xhline{1.0\arrayrulewidth}
\hline
$\mathcal{L}_{\mathrm{GLD}}$ & $\mathcal{L}_{\mathrm{LLD}}$ & $\mathcal{L}_{\mathrm{GGD}}$ & \begin{tabular}[c]{@{}c@{}}Segmentation \\ mIoU\end{tabular} & \begin{tabular}[c]{@{}c@{}}Detection \\ $AP_{50}$ \end{tabular} \\ 
\hline\hline
 &  &  & 34.28 & 31.40 \\ 
$\cmark$ &  &  & 37.65 & 45.98 \\ 
$\cmark$ & $\cmark$ &  & 37.89 & 45.39 \\ 
\rowcolor{palegreen} $\cmark$ & $\cmark$ & $\cmark$ & \textbf{39.34} & \textbf{46.38} \\ 
\Xhline{1.0\arrayrulewidth}
\hline
\end{tabular}
}
\end{minipage}

%% file: tables/ablation_3_weight_contrastive.tex
\noindent\begin{minipage}{\linewidth}
\centering
\resizebox{\textwidth}{!}{%
\begin{tabular}{cc|c|c} 
\Xhline{1.0\arrayrulewidth}
\hline
Type & Weighted & \begin{tabular}[c]{@{}c@{}}Segmentation \\mIoU\end{tabular} & \begin{tabular}[c]{@{}c@{}}Detection \\ $AP_{50}$ \end{tabular} \\ 
\hline\hline
cosine & $\xmark$ & 36.98 & 45.77 \\ 
cosine  & $\cmark$ & 35.80 & \textbf{46.01} \\ 
contrastive & $\xmark$ & 37.62 & 45.88 \\
\rowcolor{palegreen} contrastive & $\cmark$ & \textbf{37.6} & 45.98 \\ 
\Xhline{1.0\arrayrulewidth}
\hline
\end{tabular}
}
\end{minipage}

%% file: tables/ablation_2_hyperparam.tex
\noindent\begin{minipage}{\linewidth}
\centering
\resizebox{\textwidth}{!}{%
\begin{tabular}{c|cccc|cc}
\Xhline{1.0\arrayrulewidth}
\hline
& {$\lambda_{\mathrm{GLD}}$} & {$\lambda_{\mathrm{LLD}}$} & {$\lambda_{\mathrm{GGD}}$} & {$\tau$} & Segmentation & Detection \\
\hline\hline
\multirow{2}{*}{$\lambda$ scale} & 1 & 1 & 1 & 1 & \textbf{42.31} & 44.57 \\
 & 1 & 0.01 & 0.01 & 1 & 39.11 & 45.87 \\
\hline
\multirow{2}{*}{$\tau$ scale} & 1 & 0.01 & 1 & 10 & 39.44 & 44.95 \\
 & 1 & 0.01 & 1 & 0.1 & 37.53 & 45.11 \\
\hline
\rowcolor{palegreen} ATAS & 1 & 0.01 & 1 & 1 & 39.34 & \textbf{46.38} \\
\Xhline{1.0\arrayrulewidth}
\hline
\end{tabular}
}
\end{minipage}

%% file: sec/5_related_work.tex
\section{Related Works}
\label{sec:relatedworks}

\subsection{Open-Vocabulary Dense Prediction}
Open-vocabulary dense prediction extends traditional tasks like object detection and segmentation by enabling models to recognize unseen categories during training. However, fully exploiting CLIP’s capabilities for dense prediction remains challenging, prompting various adaptations in recent studies. Some studies~\cite{zsseg, zegformer, ovseg, freeseg, maft} leverage a two-stage framework, where mask proposals are first generated through an external model and then classified using CLIP. Meanwhile, weakly-supervised methods~\cite{tcl, segclip, viewco, catseg} have also been explored. Other approaches~\cite{maskclip, clipsurgery, sclip} directly adapt CLIP for dense prediction by making subtle modifications to its ViT architecture to enhance localization capabilities.

Further research has aimed to enhance CLIP’s performance by refining the alignment of its internal representations. For instance, PACL~\cite{pacl} aligns local patch tokens with text CLS tokens, while CLIPSelf~\cite{clipself} employs self-distillation to align patch tokens with the image’s global CLS token. The former, however, is fundamentally limited by its reliance on paired image-text data. Meanwhile, the latter degrades the CLIP model’s intrinsic coherence, thereby limiting its effectiveness in dense prediction tasks.

In this paper, we propose a novel self-distillation method that improves fine-grained alignment and leverages CLIP’s semantic coherence for improved performance in open-vocabulary dense prediction.

\subsection{Self-Distillation}
Self-distillation is a training method where a student model, having the same architecture as a teacher model, learns from the teacher’s knowledge. In this setup, the student model can outperform the teacher model without requiring additional networks or architectures~\cite{revisit-selfdistil, bornagain, moredtolerant}. 

There are several strategies for applying self-distillation. One involves distorting the input data and aligning the features extracted from these distorted inputs~\cite{ddgsd, lee2019rethinking}. Another method distills knowledge from the intermediate layers of the teacher model to the student model~\cite{byot, sad, sdmae}. The other approach focuses on aligning the features produced by the student model with those produced by the teacher model~\cite{robustcm, 2023maskclip, silc, clipself}. 


In this paper, we present a self-distillation approach that transfers knowledge from a teacher CLIP model by combining multi-level objectives, from global-to-local, global-to-global, and local-to-local. This method is specifically designed to enhance two key properties for dense prediction: fine-grained alignment and semantic coherence.

\subsection{Mosaic Augmentation of Object-Centric Images}
For dense prediction tasks, scene-centric datasets (e.g., the COCO dataset~\cite{coco}) are typically leveraged due to their diverse scenes and rich object instances. In contrast, several object detection methods generate a single composite image by stitching together multiple images~\cite{mosaicos, mosrep, clim}. MosRep~\cite{mosrep} employs this technique for self-supervised learning, leading to improved visual representations and enhanced performance in dense prediction tasks. Similarly, MOSAICOS~\cite{mosaicos} leverages object-centric datasets to tackle challenges in long-tail object detection by applying mosaic augmentation to group object-centric images, thereby increasing the visibility and frequency of rare objects in the training data. CLIM~\cite{clim} also adopts similar augmentation methods, albeit with the goal of generating pseudo regions of objects.


Unlike previous approaches, our work further explores the use of object-centric datasets within a distillation framework. Specifically, the mosaic augmentation employed in ATAS is designed to resolve the ambiguity of region-level representations in scene-centric images, thereby enhancing performance in open-vocabulary dense prediction tasks.


%% file: sec/6_conclusion.tex
\section{Conclusion}
\label{sec:conclusion}

In this work, we addressed the challenges in adapting CLIP for open-vocabulary dense prediction tasks, with a focus on two critical factors: semantic coherence and fine-grained alignment. Our analysis revealed that existing methods, which enhance fine-grained alignment, often compromise semantic coherence—both of which are essential for achieving high performance in tasks that demand detailed, region-level comprehension. To overcome this limitation, we proposed Any-to-Any Self-Distillation (ATAS), a novel framework that simultaneously improves both semantic coherence and fine-grained alignment using only unlabeled images. ATAS achieves substantial performance improvements across various benchmarks, including open-vocabulary object detection, and semantic segmentation, demonstrating its robustness as a solution for open-vocabulary dense prediction tasks. 

\paragraph{Acknowledgements}
This work was supported by the National Research Foundation of Korea (NRF) grant (No. RS-2024-00345809, Research on AI Robustness Against Distribution Shift in Real-World Scenarios), the Institute of Information \& communications Technology Planning \& Evaluation (IITP) grant (No. RS-2023-00232046, Development of Cloud-Based Cause Analysis Technology by Transmission of Abnormal Driving Data, RS-2025-02263754, Human-Centric Embodied AI Agents with Autonomous Decision-Making), both funded by the Korea government (MSIT).
